\documentclass[sigconf]{acmart}
\usepackage[acronym]{glossaries}
\makeglossaries
\newacronym{CNN-LSTM}{CNN-LSTM}{Convolutional Long Short Term Memory}
\newacronym{TCN}{TCN}{Temporal Convolutional Networks}

\AtBeginDocument{%
  \providecommand\BibTeX{{%
    \normalfont B\kern-0.5em{\scshape i\kern-0.25em b}\kern-0.8em\TeX}}}

\begin{document}

\title{Personalized breath based biometric authentication with wearable multimodality}

\author{Manh-Ha Bui}
\authornote{Correspondence to Manh-Ha Bui: <\texttt{hb.buimanhha@gmail.com}>.}
\affiliation{%
  \institution{VinAI Research}
  \country{Vietnam}}

\author{Viet-Anh Tran}
\affiliation{%
  \institution{Deezer Research}
  \country{France}}

\author{Cuong Pham}
\affiliation{%
 \institution{Posts and Telecommunications Institute of Technology}
 \country{Vietnam}}

\begin{abstract}
Breath with nose sound features has been shown as a potential biometric in personal identification and verification. In this paper, we show that information that comes from other modalities captured by motion sensors on the chest in addition to audio features could further improve the performance. Our work is composed of three main contributions: hardware creation, dataset publication, and proposed multimodal models. To be more specific, we design new hardware which consists of an acoustic sensor to collect audio features from the nose, as well as an accelerometer and gyroscope to collect movement on the chest as a result of an individual’s breathing. Using this hardware, we publish a collected dataset from a number of sessions from different volunteers, each session includes three common gestures: normal, deep, and strong breathing. Finally, we experiment with two multimodal models based on \acrfull{CNN-LSTM} and \acrfull{TCN} architectures. The results demonstrate the suitability of our new hardware for both verification and identification tasks.

\end{abstract}



\maketitle

\section{Introduction}
Breath is not only a vital sign for humans, but it can also be a unique characterization for individuals as breathing contains many personal chemical and physical features. Breath’s chemical features are composed of nitrogen, oxygen, carbon dioxide, water vapor, argon, etc. The volume of the chemical components is often different from person to person~\cite{fasola19repeatability,wang2015human}, while breath’s physical components include the lungs, diaphragm, intercostal muscles, bronchi, trachea, larynx, vocal tract and mouth cavity~\cite{sinues2013metabolic,zordan2006easy,korenbaum1997acoustic}. Capturing breath features is an important task not only for diagnosing \& monitoring respiration diseases, but also in terms of biometric signatures~\cite{jagmohan2017breathprint,chauhan2018performance,chauhan2017breathrnnet}. In addition, breath as a biometric authentication schema has attracted the research community as it has a wide range of potential authentication applications for the Internet of Things (IoT) and mobile devices~\cite{chauhan2018performance}, while being easy and comfortable-in-use for the users.

However, capturing all (chemical and physical) signals of breath is still a challenging task because of hardware cost and limitations. The hardware device capturing chemical signals is uncommon, this leads to significant attempts to capture breath’s physical features such as sounds of inhalation and exhalation of the breath. The majority of previous studies focus on capturing a single modality of breath’s physical features (i.e. breath sounds). Therefore, in this study, we prototype a tiny wearable IoT device that is integrated with the accelerometer, gyroscope and acoustic sensors. Our proposed wearable device is a multi-sensing modality as it can capture breath sounds as well as the movement of a user’s chest area while he/she is breathing, so this would provide richer information of breath, which could improve the performance of breath based biometric authentication systems. Moreover, we propose deep multimodal models for analyzing personalized breaths for biometric identification and verification tasks. The main contributions of this paper are made as follows.
 
\begin{itemize}
    \item \textit{Wearable device}: we design and prototype a new wearable IoT hardware device embedded with an accelerometer, a gyroscope, and an acoustic sensor. The device is tiny, light-weight, and easily mounted on the user’s chest. While the chest movements caused by lungs, diaphragm, and intercostal muscles can be captured by the accelerometer and gyroscope sensors, the sounds of bronchi, trachea, larynx and vocal tract from the nose can be captured by the acoustic sensor incorporated in the device.
    \item \textit{Dataset}: we collect a dataset from 20 subjects wearing our wearable device and perform normal, deep, and strong breaths for several sessions over one month. The collected data is then annotated using the Audacity tool. The dataset and annotations will be made available to the research community.
    \item \textit{Deep models for personalized breath identification and verification tasks}: we propose deep multimodal 1-D \acrfull{CNN-LSTM} and \acrfull{TCN} models for analyzing heterogeneous data from multiple signal streams. The models are trained with the dataset and then used for the identification task as an embedding network in the verification task. The proposed models are rigorously evaluated with two empirical experiments: monomodality and multimodality.
\end{itemize}
All source code and dataset in this work are made public on GitHub \href{https://github.com/manhhabui/personalized-breath}{\textit{https://github.com/manhhabui/personalized-breath}}.

\section{Related work}
Biometric authentication has attracted significant interest from the security research community, as its security process utilizes the unique biological characteristics of an individual to verify legal users while being effective in rejecting imposters. Biometric data such as voice, heartbeats, gesture, and motion can be captured using acoustic sensors, electrocardiograph (EEG) sensors, touch screens, and accelerometers, while common physical biometrics such as fingerprints, hand or palm geometry, and retina, iris, or facial characteristics can be captured with digital cameras~\cite{schroff201facenet,zhu2009signature,jain1997fingerprints}. Work by Hong, L. et al.~\cite{lu2011speakersense}, for example, identifies the speaker with high accuracy using mobiles with efficient energy consumption. Several works~\cite{ehatishamulhaq2018continuous,sym2016} utilized motion sensors including gravity, accelerometer, gyroscope, and magnetometer for capturing human activity patterns for biometric signatures. Touch gestures and keystroke dynamics are commonly used in human-machine interaction, utilizing the difference in touch position, the drift of the hand, and how a person presses on the screen for user identification or verification~\cite{kambourakis2014touchstroke,aviv2010smudge}. Other work such as~\cite{falconi2016ecg}, exploited electrocardiogram sensors (EEG) to record and measure the electrical potential generated by the heart and has shown EEG signals can potentially be a biometric feature for user identification.

As one of the unique biological characteristics of one individual, personalized breath was investigated for a complementary biometric identification in several research works. Electronic nose (E-nose) technology~\cite{wang2015human}, for instance, investigated 12 gas sensors consisting of six types of doped tin dioxide (SnO$_2$) sensors and one type of tungsten trioxide (WO$_3$) sensor, for user identification. This study showed that a subset of these features could identify 10 individuals with high accuracy. In addition, works that utilized acoustic sensors for biometric authentication based on breath such as BreathID~\cite{lu2020i} and Breath-Print~\cite{jagmohan2017breathprint} have shown promising results for verification and identification. The methods of~\cite{lu2020i,jagmohan2017breathprint} included the segmentation phase and frequency threshold analysis to detect people’s breathing in a period, and Gaussian mixture models were proposed for authentication.

With the advantages of deep learning for many pattern detection and recognition tasks, recent works investigated deep neural networks for analyzing breath sounds for biometric authentication. Work by CMU researchers~\cite{zhao2017speaker}, for example, proposed a \acrshort{CNN-LSTM} model for analyzing the personal sounds produced during intra-speech inhalation and used them to identify occupants in a room with 91.3\% accuracy. Other studies such as BreathRNNet~\cite{chauhan2017breathrnnet}, an extension version from BreathPrint~\cite{chauhan2018performance}, proposed neural network models for the investigation of how the models can be deployable on the devices with limited constraints. They proposed Long-Short Term Memory models (LSTM) which were mitigated using the Linear Algebra factorization technique (Singular Value Decomposition) and demonstrated the feasibility of the model when deployed on the Internet of Things devices while obtaining reasonable accuracy in human identification.

Inspired by the success of multimodality for pattern recognition tasks, combining personal biological characteristics captured from multiple different sensors has been investigated for biometric authentication. For example, work by Al-Waisy et. al.~\cite{alwaisy2017multimodal} combined image features from face with left and right iris captured from the digital camera and human voice captured from a microphone~\cite{damousis2012four,poh2004hybrid} to enhance the authentication accuracy. In addition, \cite{barbu2015multimodal} fused features from the human face, speech, and iris to effectively verify the users, while~\cite{hammad2018multimodal} combined fingerprint images and ECG data captured from the electrocardiograph sensor into a unified framework to produce a high performance of the authentication systems. In general, literature works have shown a significant improvement in the performance of biometric authentication-based multimodality. This motivates us to investigate the effectiveness of multimodality for user identification and verification tasks using personal breath. Our work is distinctive from other works as we combine breath sounds and chest movement captured from heterogeneous sensors embedded into an IoT wearable device, which have not yet been investigated in the past.

\section{Hardware}
\begin{figure}[ht!]
\begin{center}
\includegraphics[width=1.0\linewidth]{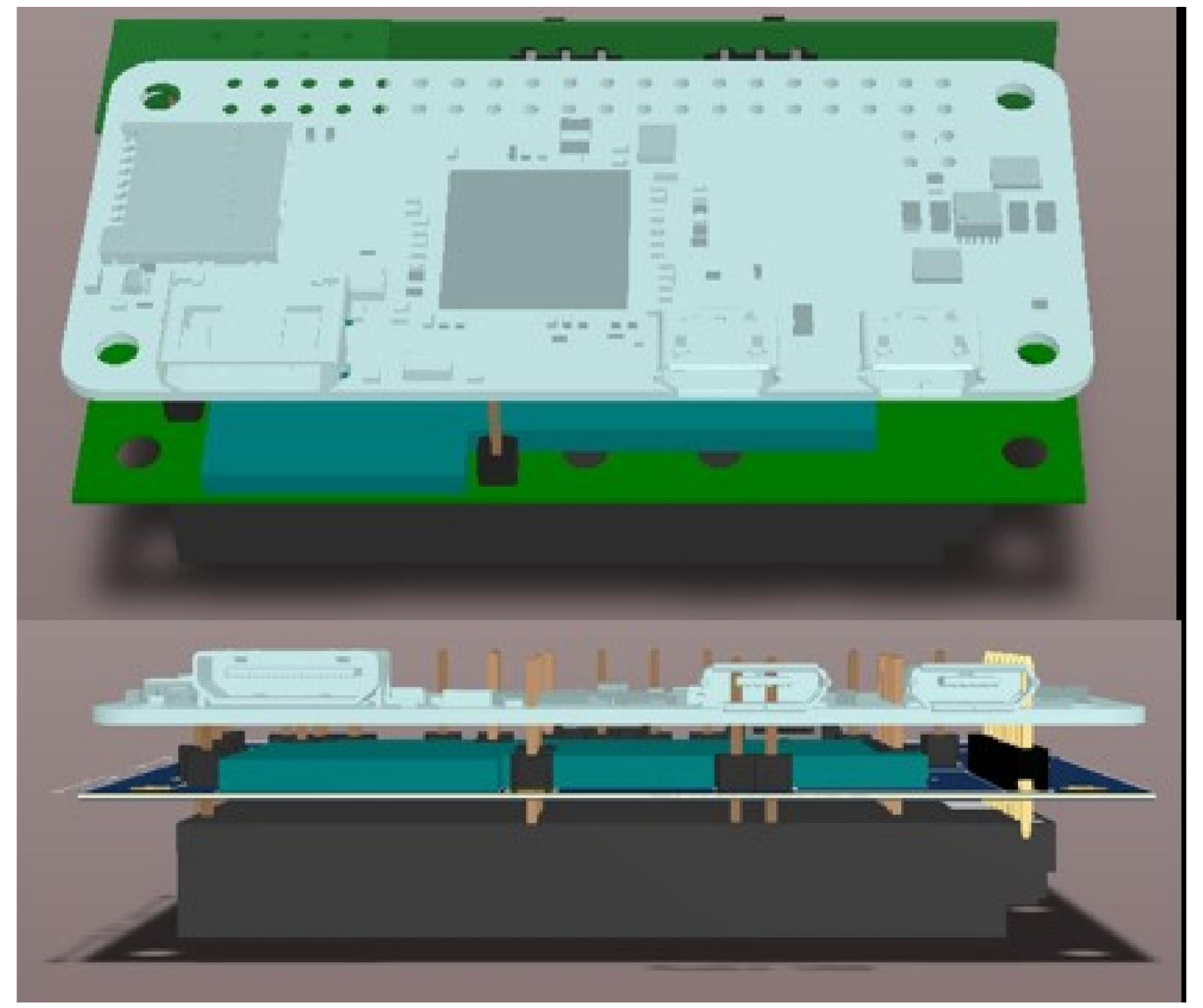}
\end{center}
   \caption{PCB wearable device design.}
\label{fig:hardware}
\end{figure}
We custom design and prototype an IoT wearable device. The device incorporates a microprocessor, wireless communication, battery, memory, and multiple sensors. Our goal is to make the size of the device miniature for ease of being worn on the user’s chest area with low power consumption. For the first version of the prototype, we employ the raspberry PI-Zero operating at 1 GHz single-core CPU with 512Mb of RAM and a built-in wireless communication module. The size of the device in the 3D dimensions is 35.8 x 65 x 9.1 mm with a weight of just 25 grams (see Figure~\ref{fig:wearer} - left). In addition, an acoustic sensor, an accelerometer, and a gyroscope is integrated into the device. The wearable device’s rechargeable Li-Polymer battery can continuously capture and pre-process the sensing signals for up to 8 hours, and for up to 7 days in hibernate mode. The sensing range between the device to IoT gateway is up to 25 meters. Figure~\ref{fig:hardware} shows the 3D PCB of the prototype design. In this study, we choose the sampling frequencies of 44.1 kHz for the acoustic sensor and of 50Hz for the accelerometer and gyroscope sensors.

\section{Dataset}
\subsection{Data collection}
\begin{figure}[ht!]
\begin{center}
\includegraphics[width=1.0\linewidth]{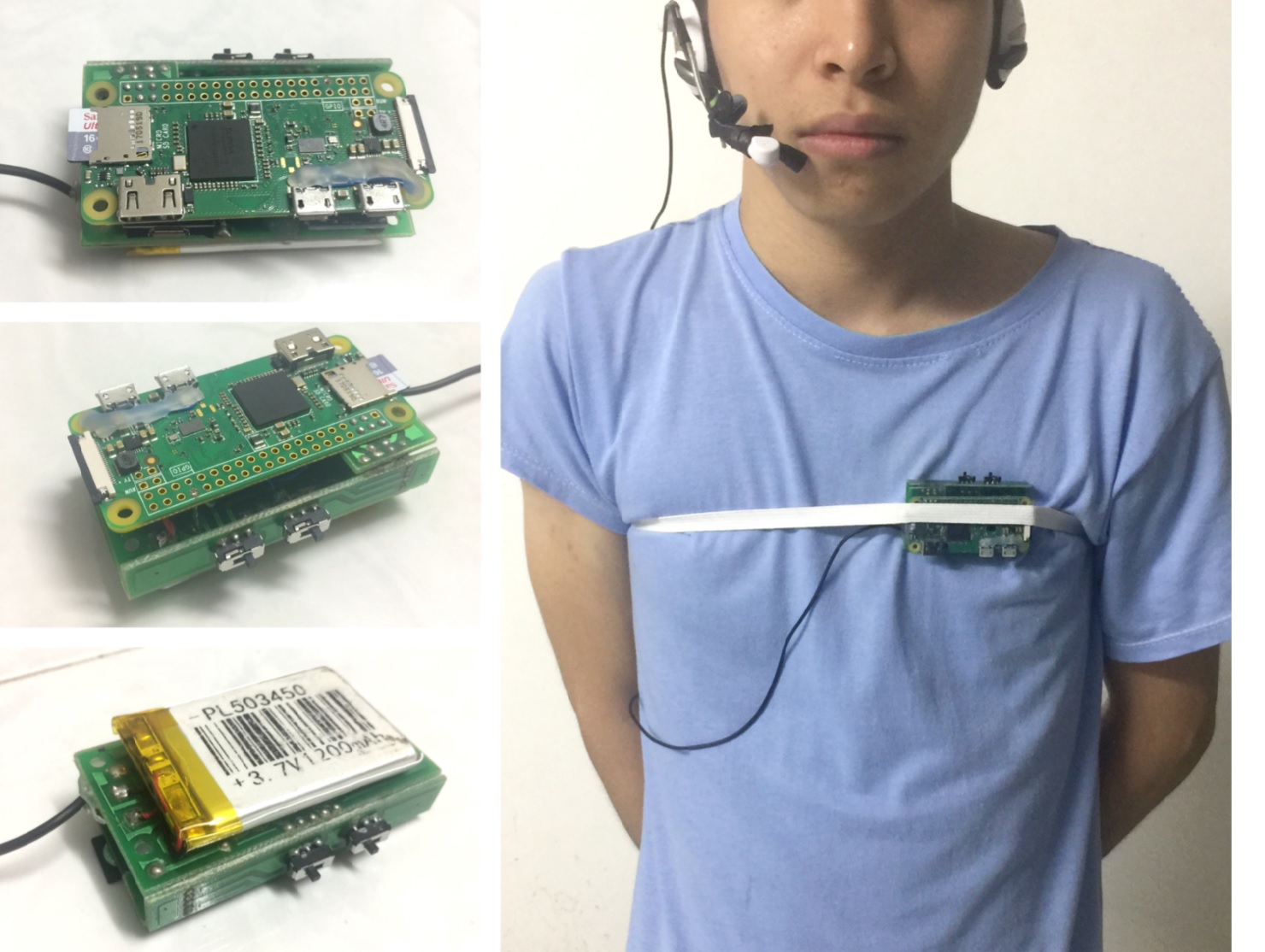}
\end{center}
   \caption{Setup hardware on subject.}
\label{fig:wearer}
\end{figure}

Basically, a breath includes two phases: inhalation and exhalation, whose variation highly depends on an individual’s physical characteristics including sex, age, weight, and health condition. These variations impact directly on amplitude, frequency, and duration of breath. To the best of our knowledge, a breath dataset has yet to be published. Therefore, in this work, we collect a breath dataset as one of our contributions. 20 subjects (4 females and 16 males) were asked to wear the device on the chest area. The microphone is mounted close to the mouth and a rubber band is worn to fasten the device on the chest for keeping it fixed (see Figure~\ref{fig:wearer}). The accelerometer and gyroscope sensors capture the chest movement, and the acoustic sensor can capture the sound signals while the subject was intentionally performing breaths on several days in one month. To generalize, the subjects were asked to collect 3 breathing instances for one session at different times of the day to cover intrapersonal variation. The subject performs 3 different types of breath for each session. It can be observed that the data is also recorded after the subject has performed daily activities such as walking, working, driving, going to work, playing sports, etc. To minimize the noise of the sensors, the subject stands stationary during the data collection session.

\subsection{Breath gesture annotation}
Annotators are given three breath labels including \textit{strong}, \textit{normal}, and \textit{deep}. Annotators are advised to listen to samples of each type of breath and they have also explained the definition of each breath label (see Figure~\ref{fig:signalvisualization}). A breath is called \textit{strong} if, on average, its duration is around 0.5 to 2 seconds (shortest when compared to normal and deep); while a \textit{normal} breath lasts long from 1.5 to 2.5 seconds and the \textit{deep} breath has the longest duration, approximately between 2 and 3.5 seconds. All three of the breath types are in a common breathing phase consisting of an inhalation and an exhalation.

\subsection{Data statistics}
\begin{figure}[ht!]
\begin{center}
\includegraphics[width=1.0\linewidth]{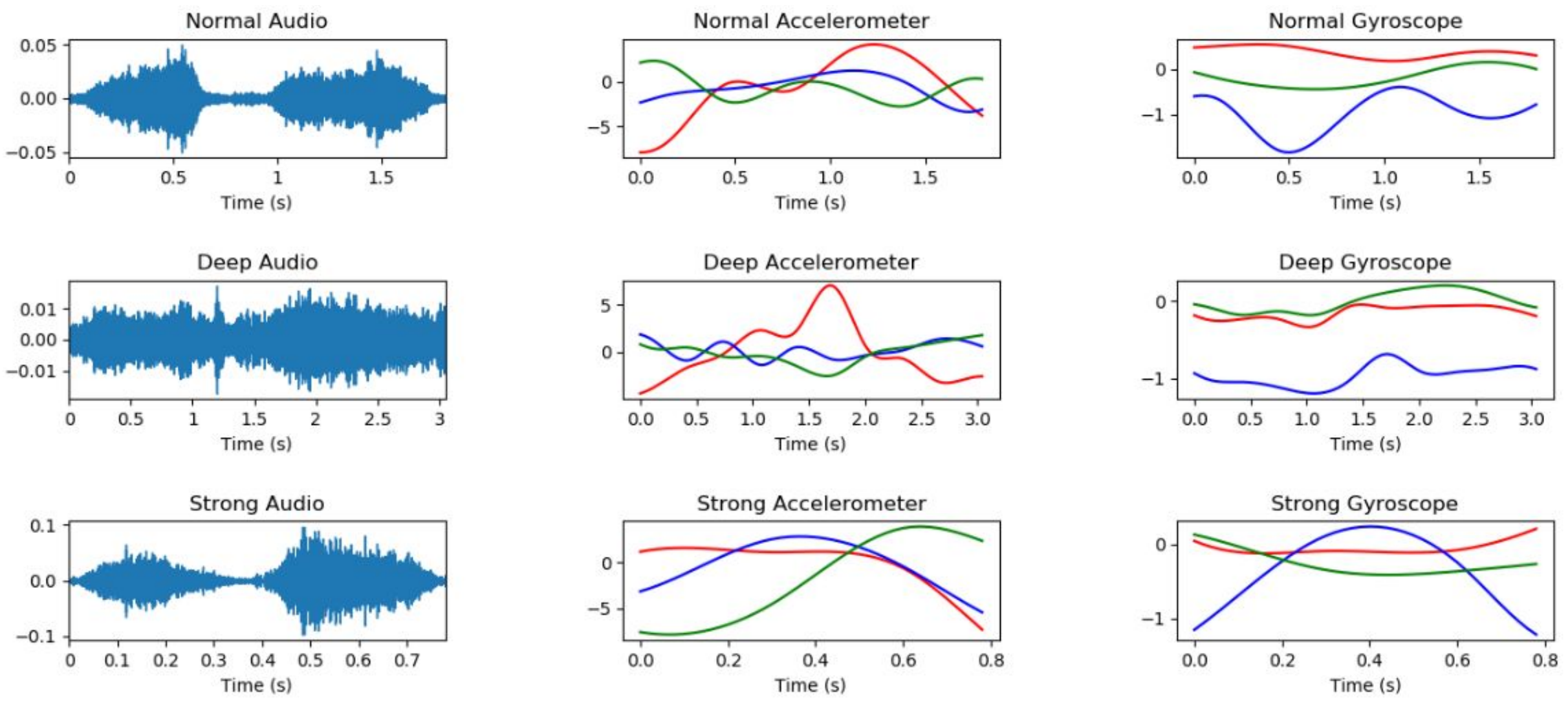}
\end{center}
   \caption{Audio, accelerometer, and gyroscope signals for each breathing type, where x, y, the z-axis of motion signals are denoted by red, blue, and green respectively.}
\label{fig:signalvisualization}
\end{figure}

Figure~\ref{fig:signalvisualization} illustrates 3 sensing signals including audio, accelerometer, and gyroscope for each type of breathing from a random subject in one session.

\begin{figure*}[ht!]
\begin{center}
\includegraphics[width=0.65\linewidth]{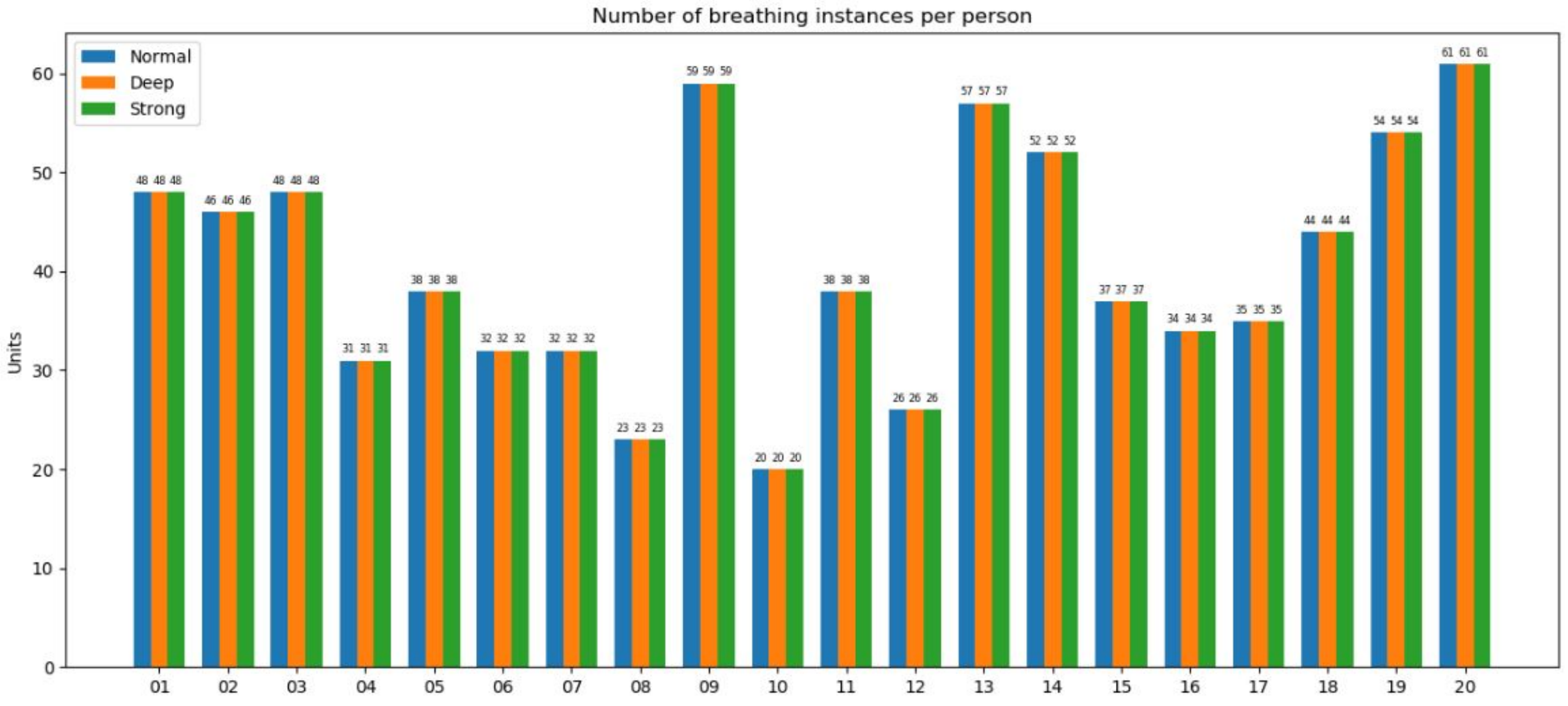}
\end{center}
   \caption{The number of breathing instances per subject person.}
\label{fig:instances}
\end{figure*}

After data collection, because of flexible time for each subject, the collected dataset is relatively unbalanced among the subject’s instances. Figure~\ref{fig:instances} shows the number of breathing instances per subject. The most instances for a subject that we could collect was 61 while the least was 20, on average this number was approximately 40 samples.

Table~\ref{tab:length} shows the min, max, median, mean, and standard deviation of the length of time from all of the instances. Overall, the duration of strong breathing gestures is the shortest, and deep is the longest. On average, normal breathing duration is around 2.12 seconds while they vary between 0.96 and 4.13 seconds. Deep breathing is a bit longer with an average of 2.58 and its variation in a range of 1.35-4.49. The shortest is the strong breathing with a mean of 1.02 and a length variation between 0.4 and 2.48.

\begin{table}[ht!]
\caption{The length time in seconds of each breathing type.}
\label{tab:length}
\centering
\begin{tabular}{lcccc}
\toprule
\textbf{Type} & \textbf{Min} & \textbf{Max} & \textbf{Median} & \textbf{Mean $\pm$ Std} \\
\midrule
Normal & 0.96 & 4.13 & 2.04 & 2.12 $\pm$ 0.47\\
Deep & 1.35 & 4.49 & 2.50 & 2.58 $\pm$ 0.61\\
Strong & 0.40 & 2.48 & 0.86 & 1.02 $\pm$ 0.46\\
\bottomrule
\end{tabular}
\end{table}

\section{Methodology}
\subsection{Data Preprocessing}
Due to time variation in breathing instances, we use zero-center padding to keep the shape of a neural network’s batch equal. Specifically, all samples have the same length of 4.5 seconds for normal and deep, while strong breath after the padding has a length of 2.5 seconds. For accelerometer and gyroscope modalities, with 50Hz sampling frequency, the raw signals are used to make use of the representation power of the neural network. On the other hand, the audio signal is first downsampled to 16kHz. 20 Mel Frequency Cepstral Coefficients (MFCC) features are then extracted from every 32ms window with a 20ms frameshift to characterize audio modality.
\subsection{Identification task} 
We describe below two proposed models for the identification task, these models allow us to determine an unknown personal identity from signals coming from both chest movements and nose sound: one inspired from \acrshort{CNN-LSTM}~\cite{zhao2017speaker}, another is full causal convolution with \acrshort{TCN} architecture~\cite{bai2018empirical}.

\begin{figure}[ht!]
\begin{center}
\includegraphics[width=1.0\linewidth]{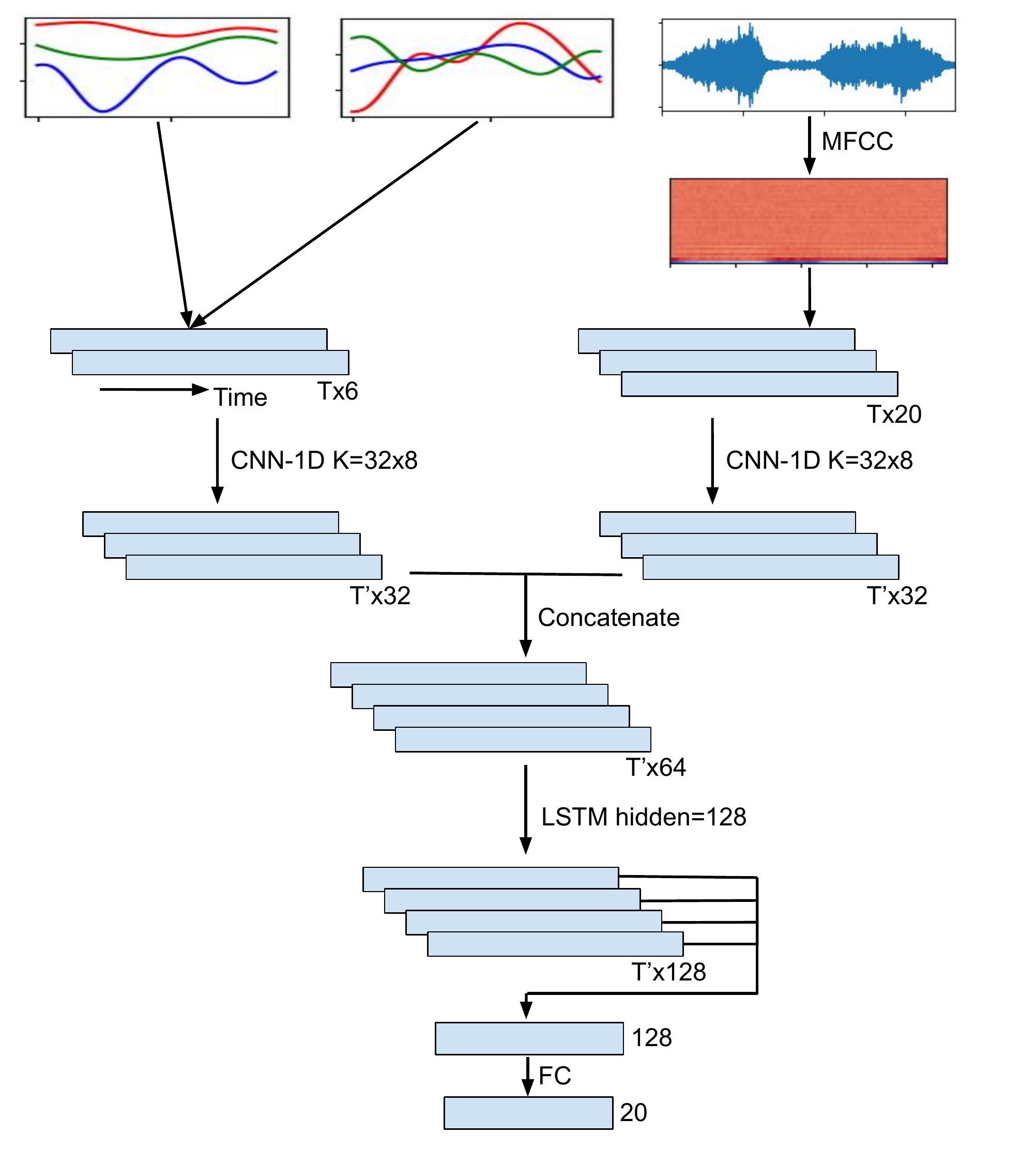}
\end{center}
   \caption{\acrshort{CNN-LSTM} MultiModality Architecture.}
\label{fig:cnnlstm}
\end{figure}

\textit{\acrshort{CNN-LSTM} Models.} Figure~\ref{fig:cnnlstm} shows multimodal architecture inspired by the recent \acrshort{CNN-LSTM} model proposed by authors in~\cite{zhao2017speaker}. One small modification in our model is the use of 1D convolution instead of 2D convolution, this is because of different semantics between the two dimensions. In addition, there is an exclusion of max-pooling because of the small number of MFCC coefficients and motion channels. Therefore, our model is composed of a 1-D convolutional layer for each modality, a fusion layer, an LSTM layer, and a fully connected layer.

For an input $X \in R^{T\times F}$ where $T$ is the number of samples in the time domain, $F$ is the number of channels in the case of chest motion signals or the number of MFCC coefficients in the case of the audio signal, the 1-D convolutional layer convolutes it to extract features with $L$ filters $W_i \in R^{k}, i = 1, ..., L$ with filter size $k$. These features are then passed through a rectified function $X_{i}^{conv} = \max(0, X * W_i)$ to obtain the sequence $X^{conv} = \left \{ X_{1}^{conv}, ..., X_{L}^{conv} \right \}$, where $X^{conv} \in R^{T_c \times L}$ with $T_c < T$. These sequences coming from both modalities are concatenated in channel dimension by a fusion layer to become $X^{fusion} \in R^{T_c \times 2 * L}$ before feeding into an LSTM layer, this layer allows us to explore the correlation of the features along the time domain by using a sequence of memory units. Each unit consists of one memory cell and three control gates (input, output and forget). Finally, the output features $h$ from the LSTM layer corresponding to the last time in a breathing period was selected, then fed into a fully connected layer to obtain the multi-class likelihood output $\hat{y} = \left \{\hat{y_1}, .., \hat{y_n}\right \}$ for n identification subjects, where $\hat{y_i} =\frac{w_i^T h}{\sum_{j=1}^{n}w_j^Th},i=1,...,n$ and $w_i$ is the weight in the fully connected layer.

\begin{figure}[ht!]
\begin{center}
\includegraphics[width=1.0\linewidth]{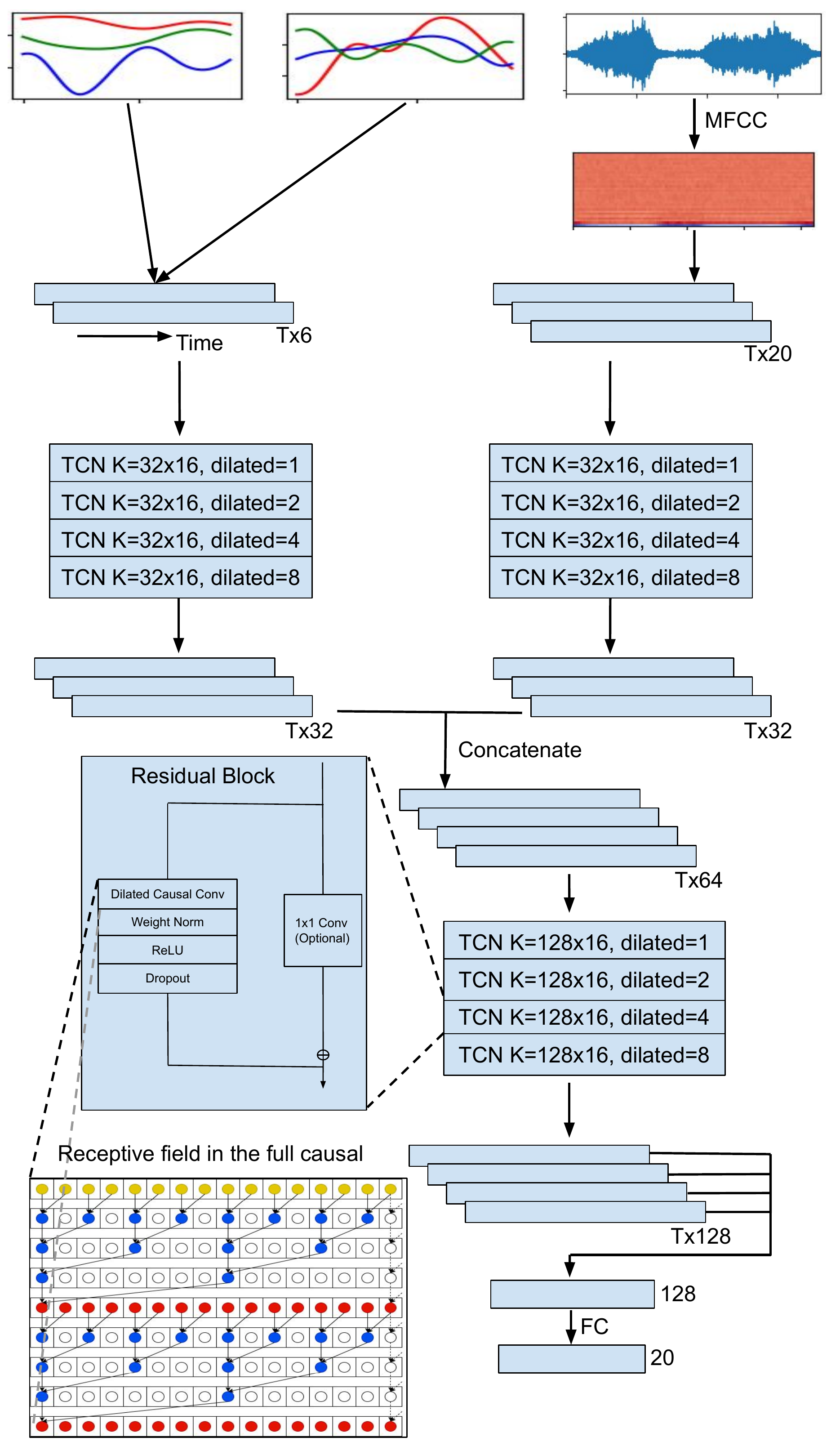}
\end{center}
   \caption{\acrshort{TCN} Multimodality Architecture.}
\label{fig:tcn}
\end{figure}

\textit{\acrshort{TCN} Models.} There are two main drawbacks of the mentioned \acrshort{CNN-LSTM} models. First, with only one convolutional layer, the feature vectors are represented at a low level which is not enough to capture representative characteristics of each modality. More importantly, the convolutional layer is non-causal and not suitable for real-time application. Another disadvantage is the LSTM layer. First, choosing a smaller filter size in the convolutional layer increases the length of the concatenated layer output, which makes the training of the LSTMs unmanageable. Second, the large number of parameters in a deep LSTM network significantly increases its computational cost and limits its applicability to low-resource, low-power platforms such as wearable devices. To tackle these drawbacks, we proposed a causal convolutional model based on \acrshort{TCN}, a recent by-default architecture for sequence modeling because of their superior advantages compared to RNNs and LSTM such as parallelizability, stable gradient, low memory requirement, and speedup for training~\cite{bai2018empirical,luo2019conv}.

Figure~\ref{fig:tcn} shows the detail of our full causal convolutional \acrshort{TCN} architecture. The non-causal convolutional layer and the LSTM layer in the \acrshort{CNN-LSTM} architecture are replaced by two causal \acrshort{TCN}s. Each \acrshort{TCN} includes four residual blocks with the dilation factor $d$ increasing exponentially $d=O(2^i), i = 0, ..., 3$ in order of blocks. Each block comprises one 1-D dilated causal convolution layer with a filter size to $k’$ weight normalization to the convolutional filters, rectified linear unit (ReLu) for non-linear activation, and spatial dropout for regularization. In addition, in each block, $1 \times 1$ convolution is applied to the residual path to ensure the input and the output tensors have the same shape. Therefore, the feature vector of the first \acrshort{TCN} for each modality is $X_i^{dconv_1} = X_i^{1 \times 1} + \max(0, X * W_i )$, where $X^{1 \times 1} \in R^{T \times L_1}$ and $L_1$ filters $W_i \in R^{k’ * d}, i = 1, ..., L_1$ with filter size $k' * d$. In the second \acrshort{TCN}, the number of filters increases to $L_2$ to capture information in higher dimensions. It maps the collection of fusion feature vectors $X^{fusion} \in R^{T \times 2 * L_1}$ to $X^{dconv_2} \in R^{T\times L_2}$ with ${L_2 > L_1}$.

\subsection{Verification task}
In the verification task, the subject claims to be of a certain identity. The biometric features extracted from the chest movement and the nose sound are used to verify this claim. To this end, we consider the verification as a downstream task of the identification procedure described above by using the last layer in the trained networks for identification to represent a personal breath vector. Each subject has a prototype vector representing their breathing characteristic which is computed from the mean of embedded vectors belonging to him/her. This representative vector can be visualized as a centroid from the subject’s vectors in the high dimensional space. During the inference, the euclidean distance between the embedded vector of the testing instance and the prototype vector of the verified subject is used to decide if the breathing instance belongs to that subject. If this distance is smaller than a threshold, the claim will be accepted, otherwise, it will be rejected. This threshold is determined by the intersection of the False Positive curve and False Negative curve. Equation~\ref{eq:1} shows our approach for the verification task:
\begin{equation}\label{eq:1}
\left\{\begin{matrix}
accept & \text{ if } d(f_w(\hat{x}),c_k) < \epsilon,\\ 
reject & \text{ otherwise,}
\end{matrix}\right.
\end{equation}
where d denotes the Euclidean distance, $f_w: R^{T\times F}\rightarrow R^m$ is an embedding model with parameters $w$ learnt in identification task, $\hat{x} \in R^{T\times F}$ is a testing instance, $\epsilon$ is the threshold such that $FPR(\epsilon)=FNR(\epsilon)$ and $c_k$ is centroid embed vector of subject $k$, calculated as follow~\ref{eq:2}:
\begin{equation}\label{eq:2}
c_k=\frac{1}{\left | S_k \right |}\sum_{x_i\in S_k}f_w(x_i),
\end{equation}
where $S_k$ denotes the set of training instances $x_i \in R^{T \times F}$ in subject $k$.

\section{Experiments}
\subsection{Experiments setting}
\textit{Dataset.} For each breath type of each person, 10 instances are selected randomly, 5 for validation and 5 others for testing. The remaining instances are kept for training. In total, 100 instances from 20 subjects are used for validation, 100 instances to test and 615 instances to train for each breathing type. In addition, two different scenarios are designed for the verification task. For the first scenario, each subject has 5 true instances and 95 false instances representing the impersonal attacks in the test set. In the second scenario, we assume there are some imposters who are outside of the database. Four subjects are randomly selected and all their instances are filtered out. Therefore, they are unseen during the training for the identification task.
 
\textit{Evaluation methodology.} The experiments are run on 100 train/test splitting with the same strategy described above. The results are then recorded by mean and standard deviation. Classification accuracy is used to evaluate the identification performance while Equal Error Rate (EER) is derived to represent the verification accuracy.
 
\textit{Parameters setting.} We train our proposed models by using back-propagation. The batch size is set to 128. The number of epochs is up to 1000. Adam optimizer is used for all models. The learning rate is 0.001, divided by half up to four times when validation categorical cross-entropy loss does not decrease before stopping.

\subsection{Identification task}
\begin{table}[ht!]
\caption{Identification accuracy with three breath types in monomodal and multimodal \acrshort{CNN-LSTM} models.}
\label{tab:cnn_lstm}
\centering
\begin{tabular}{lccc}
\toprule
\textbf{Type} & \textbf{Acce+Gyro} & \textbf{Audio} & \textbf{Multimodality}\\
\midrule
Normal & 83.13\% $\pm$ 4.46 & 93.22\% $\pm$ 2.60 & \textbf{97.06\% $\pm$ 1.24}\\
Deep & 81.00\% $\pm$ 5.00 & 95.72\% $\pm$ 1.58 & 96.88\% $\pm$ 1.55\\
Strong & 62.34\% $\pm$ 6.3 & 96.23\% $\pm$ 1.86 & 96.7\% $\pm$ 1.71\\
\bottomrule
\end{tabular}
\end{table}
Table~\ref{tab:cnn_lstm} shows the identification accuracy of monomodal and multimodal models based on \acrshort{CNN-LSTM} architecture in three types of breath. Our multimodal model confronts two baselines, each respectively taking the chest movement (accelerometer \& gyroscope) and the audio as input. We highlight the best results which have the p-value $\leq 0.05$ in significant tests. The results reveal the multimodality has outperformed monomodal models in all types of breath. Especially in normal breath, the multimodality reaches 97\%, higher than audio around 4\%, and significantly higher than chest movement 14\% in absolute change. Similarly in deep and strong breaths, although the accuracy for multimodality is lower than the normal breath of about 0.5\%, it still outperforms two monomodal models in each breath type. Moreover, the results from the audio modality prove that it contains valuable biometric information for personal identification, as found in previous works~\cite{lu2020i,jagmohan2017breathprint,chauhan2018performance,chauhan2017breathrnnet}, especially in the deep and strong breath where the accuracies are above of 95.5\%. In contrast, the system with chest movement as input has the lowest performance because of noise appearing in the person’s movements and less representative information compared to audio. However, it still impacts positively on personal identification as a complementary modality for audio. In addition, we observe that the duration in strong breath is significantly shorter than normal and deep breath instances, this leads to a lack of movement features of each identity in the authentication task. As a result, there is a drop of about 20\% in chest movement’s accuracy in the strong breath when compared with others’ breath.

To confirm our intuition on the useful contribution of the chest movements in personal identification, we continue to compare the performance of monomodal and multimodal causal \acrshort{TCN} in table~\ref{tab:tcn}. The results show that even in the full causal architecture, the performance of the multimodal system is still significantly better than the monomodal one. Indeed, the identification accuracy of multimodality in normal breath is still the highest, achieving 95.20\% while that of audio and motion monomodality only stayed at 90.57\% and 85.52\%. Similar trends are observed in deep and strong breaths, the multimodality for both reaches around 94\%. Compared with the \acrshort{CNN-LSTM} system, \acrshort{TCN} is a little inferior because of the causal constraint. It suggests that \acrshort{TCN} is a competitive candidate for real-time applications on wearable devices.
 
\begin{table}[ht!]
\caption{Identification accuracy with three breath types in monomodal and multimodal \acrshort{TCN} models.}
\label{tab:tcn}
\centering
\begin{tabular}{lccc}
\toprule
\textbf{Type} & \textbf{Acce+Gyro} & \textbf{Audio} & \textbf{Multimodality}\\
\midrule
Normal & 85.52\% $\pm$ 3.59 & 90.57\% $\pm$ 2.89 & \textbf{95.20\% $\pm$ 2.17}\\
Deep & 84.56\% $\pm$ 4.05 & 91.93\% $\pm$ 2.71 & 93.75\% $\pm$ 2.52\\
Strong & 74.93\% $\pm$ 4.62 & 93.39\% $\pm$ 2.28 & 94.00\% $\pm$ 2.54\\
\bottomrule
\end{tabular}
\end{table}

\subsection{Verification task}
\begin{figure*}[ht!]
\begin{center}
\includegraphics[width=0.9\linewidth]{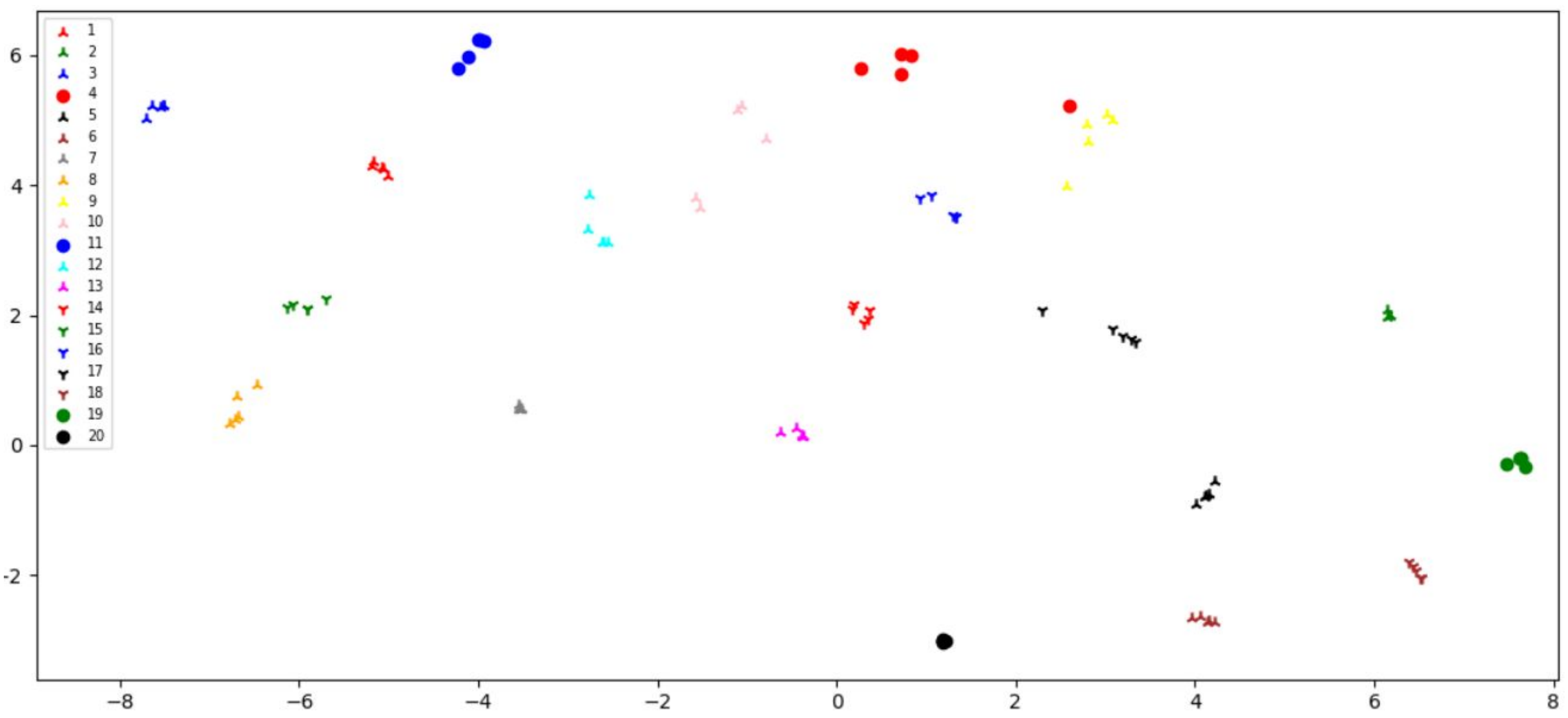}
\end{center}
   \caption{Test set projected in 2-dimensional space by using trained \acrshort{CNN-LSTM} multimodality from identification task in normal breath.}
\label{fig:tsne}
\end{figure*}

Figure~\ref{fig:tsne} visualize the embedding space of test instances, extracted from the last layer of \acrshort{CNN-LSTM} for the identification task and projected in two dimensions using t-distributed Stochastic Neighbor Embedding (tSNE)~\cite{laurens2008tsne}. Circle dots and triangle dots denote female and male subjects respectively. Instances coming from the same subject tend to close together while instances from different subjects are far apart. This property lets us leverage the trained identification models to handle the downstream verification task.

Table~\ref{tab:eer_cnnlstm} shows verification performances in EER by using the last layer of \acrshort{CNN-LSTM}. The results are related to identification accuracy. The multimodality still has a better performance compared to monomodality. In addition, the normal breath is still the best of the three types when compared with other breaths in multimodality. In the first scenario, when combining multiple signals, the EER for deep and strong breath is around 1,8\%, smaller than only using one single signal. The performance is even better in normal breath with multimodality which achieves the lowest percentage of 1.4\% while these rates of monomodality in audio and motion signals are much higher, at about 3\% and 6\% respectively. In the second scenario, although the performance is worse than the first because of more false instances from imposters who are outside the training dataset, the EER is still admissible in all of the three breath types with model fusion, under 2.4\% in all types and especially in normal breath with only 1.67\%.

\begin{table}[ht!]
\vskip -0.1in
\caption{Equal Error Rate (EER) with three breath types in monomodality and multimodality \acrshort{CNN-LSTM} embedding models in two scenarios.}
\label{tab:eer_cnnlstm}
\centering
\begin{tabular}{lccc}
\toprule
\textbf{Type} & \textbf{Acce+Gyro} & \textbf{Audio} & \textbf{Multimodality}\\
\cline{1-4}\multicolumn{4}{c}{First scenario}  \\
\midrule
Normal & 6.24\% $\pm$ 1.62 & 3.00\% $\pm$ 1.29 & \textbf{1.41\% $\pm$ 0.85}\\
Deep & 8.12\% $\pm$ 2.19 & 2.56\% $\pm$ 1.25 & 1.76\% $\pm$ 0.98\\
Strong & 12.03\% $\pm$ 2.37 & 2.01\% $\pm$ 1.09 & 1.84\% $\pm$ 1.10\\
\cline{1-4}\multicolumn{4}{c}{Second scenario}  \\
\midrule
Normal & 6.20\% $\pm$ 1.75 & 3.15\% $\pm$ 1.32 & \textbf{1.67\% $\pm$ 0.94}\\
Deep & 8.85\% $\pm$ 2.37 & 2.64\% $\pm$ 1.29 & 2.36\% $\pm$ 1.17\\
Strong & 13.05\% $\pm$ 2.52 & 2.20\% $\pm$ 0.98 & 2.00\% $\pm$ 1.01\\
\bottomrule
\end{tabular}
\vskip -0.1in
\end{table}

Regarding \acrshort{TCN} models, table~\ref{tab:eer_tcn} shows that the results in EER are higher than \acrshort{CNN-LSTM} because of obvious reasons in causal and non-causal convolution. However, when using model fusion, the multimodality continues to perform better than using only one single signal. The multimodality for normal breath achieved the lowest EER in the first scenario with only 2.54\% while audio and motion monomodality is much higher than around 3\% and 5\%. The EER in multimodality of deep and strong are around 4.1\%, higher than in normal breath; however, when compared with monomodality models in each type, these results are still better than remarkable. Similarly in the second scenario, normal breath achieved 3.37\% while that of deep and normal are 4.89\% and 4.46\% respectively, these results are still lower than using accelerometer \& gyroscopes or audio signals individually.
 
\begin{table}[ht!]
\vskip -0.1in
\caption{Equal Error Rate (EER) with three breath types in monomodal and multimodal \acrshort{TCN} embedding models in two scenarios.}
\label{tab:eer_tcn}
\centering
\begin{tabular}{lccc}
\toprule
\textbf{Type} & \textbf{Acce+Gyro} & \textbf{Audio} & \textbf{Multimodality}\\
\cline{1-4}\multicolumn{4}{c}{First scenario}  \\
\midrule
Normal & 7.03\% $\pm$ 1.95 & 5.72\% $\pm$ 1.97 & \textbf{2.54\% $\pm$ 1.06}\\
Deep & 11.25\% $\pm$ 2.04 & 5.30\% $\pm$ 1.71 & 4.18\% $\pm$ 1.53\\
Strong & 12.51\% $\pm$ 2.09 & 5.17\% $\pm$ 1.51 & 4.05\% $\pm$ 1.42\\
\cline{1-4}\multicolumn{4}{c}{Second scenario}  \\
\midrule
Normal & 8.32\% $\pm$ 2.24 & 6.31\% $\pm$ 2.10 & \textbf{3.37\% $\pm$ 1.22}\\
Deep & 11.44\% $\pm$ 2.43 & 5.86\% $\pm$ 1.87 & 4.89\% $\pm$ 1.64\\
Strong & 13.65\% $\pm$ 2.58 & 5.21\% $\pm$ 1.62 & 4.46\% $\pm$ 1.41\\
\bottomrule
\end{tabular}
\vskip -0.1in
\end{table}

\section{Conclusion}
We design and prototype an IoT wearable device for capturing breath sounds and chest movement signals of personalized breath gestures. In addition, we propose two deep models \acrshort{CNN-LSTM} and \acrshort{TCN} for analyzing heterogeneous data for the identification and verification tasks. Our experiments have demonstrated that our proposed multimodal model has outperformed the monomodal model in terms of identification accuracy and equal error rate (EER) in three types of breath, in which multimodality identification accuracy achieves around 96.8\% for \acrshort{CNN-LSTM} and 94\% for \acrshort{TCN}. In addition, these models are used as embedding networks for the verification task and achieve under 5\% EER in all evaluation scenarios. The results also prove that normal breath is the most suitable type to use in the biometric domain because of the highest accuracy in identification and lowest EER in the verification task. Future work will improve the wearable device as well as the fusion models applicable for real-world multimedia biometric authentication applications. We hope that our dataset and the proposed models that we develop in this paper will facilitate fundamental progress in understanding the behavior of both accelerometer \& gyroscope and acoustic signals in personalized breath gestures.

\bibliographystyle{ACM-Reference-Format}
\bibliography{sample-base}

\end{document}